%% file: lapq_eccv.tex
\documentclass[runningheads]{llncs}
\pdfoutput=1 % arxiv

\usepackage[dvipsnames]{xcolor} 
\usepackage{graphicx}
\usepackage{comment}
\usepackage{amsmath,amssymb} % define this before the line numbering.
\usepackage{color}
\usepackage{ulem}

\usepackage{epsfig}
\usepackage{graphicx}
\usepackage{amsmath}
\usepackage{amssymb}

\usepackage{cellspace}
\usepackage{amsfonts}
\usepackage{physics} 
\usepackage[protrusion=true,tracking=true,kerning=true,expansion,final]{microtype}      % microtypography
\usepackage[dvipsnames]{xcolor} %colors
\usepackage{booktabs} % for professional tables
\usepackage{subcaption}
\usepackage{nicefrac} % nice flat fraction
\usepackage{cancel}
\usepackage[numbers,sort&compress]{natbib}
\usepackage{xspace}

\usepackage{gensymb}
\usepackage{bbding}
\usepackage{mathtools}
\usepackage{centernot} % not parallel, etc.
\usepackage{bigints}
\usepackage{dsfont} %mathbb 1
\usepackage{esint} % beatiful integrals
\usepackage{authblk}
\usepackage{multirow}
\usepackage{enumitem}

\usepackage{algorithm}
\usepackage{algpseudocode}
\usepackage{algorithmicx}

\usepackage{varioref} % references to show page
\usepackage[pagebackref=true,breaklinks=true,letterpaper=true,colorlinks,bookmarks=false]{hyperref}
\usepackage[capitalise]{cleveref} 
\crefname{appsec}{appendix}{appendices}
\Crefname{appsec}{Appendix}{Appendices}

\usepackage{url}

\definecolor{mydarkblue}{rgb}{0,0.08,0.45}
\definecolor{urlcolor}{rgb}{0,.145,.698}
\definecolor{linkcolor}{rgb}{.71,0.21,0.01}
\hypersetup{ %
	pdftitle={Loss Aware Post-training Quantization}, 
	pdfauthor={Yury Nahshan, Brian Chmiel, Chaim Baskin, Evgenii Zheltonozhskii, Ron Banner, Alex M. Bronstein, and Avi Mendelson},
	pdfsubject={},
	pdfkeywords={},
	pdfborder=0 0 0,
	pdfpagemode=UseNone,
	breaklinks=true,  % so long urls are correctly broken across lines
	colorlinks=true,
	letterpaper=true,
	bookmarks=false,
	urlcolor=urlcolor,
	linkcolor=linkcolor,
	citecolor=mydarkblue,
	filecolor=mydarkblue,
	pdfview=FitH}

\renewcommand*{\backref}[1]{} % for backref < 1.33 necessary
\renewcommand*{\backrefalt}[4]{%
	\ifcase #1 %
	\or
	(cited on p. #2)%
	\else
	(cited on pp. #2)%
	\fi
}
\makeatletter
\renewcommand{\@biblabel}[1]{#1.}
\makeatother

\vbadness=\maxdimen
\usepackage{bm}
\renewcommand{\vb}{\bm}

\usepackage{docmute} 
\usepackage{xr}

\begin{document}
% \renewcommand\thelinenumber{\color[rgb]{0.2,0.5,0.8}\normalfont\sffamily\scriptsize\arabic{linenumber}\color[rgb]{0,0,0}}
% \renewcommand\makeLineNumber {\hss\thelinenumber\ \hspace{6mm} \rlap{\hskip\textwidth\ \hspace{6.5mm}\thelinenumber}}
% \linenumbers
\pagestyle{headings}
\mainmatter
\def\ECCVSubNumber{7380}  % Insert your submission number here

\title{Loss Aware Post-training Quantization} % Replace with your title

% INITIAL SUBMISSION 
\begin{comment}
\titlerunning{ECCV-20 submission ID \ECCVSubNumber} 
\authorrunning{ECCV-20 submission ID \ECCVSubNumber} 
\author{Anonymous ECCV submission}
\institute{Paper ID \ECCVSubNumber}
\end{comment}
%******************

% CAMERA READY SUBMISSION
%\begin{comment}
\titlerunning{Loss Aware Post-Training Quantization}
% If the paper title is too long for the running head, you can set
% an abbreviated paper title here
%

\newcommand*\samethanks[1][\value{footnote}]{\footnotemark[#1]}

\author{Yury Nahshan\thanks{Equal contribution.}\inst{1} \and Brian Chmiel\samethanks[1]\inst{1,2} \and Chaim Baskin\samethanks[1]\inst{2} \and Evgenii Zheltonozhskii\samethanks[1]\inst{2}\orcidID{0000-0002-5400-9321} \and Ron Banner\inst{1} \and Alex M. Bronstein\inst{2} \and Avi Mendelson\inst{2}}
\authorrunning{Y. Nahshan et al.}
% First names are abbreviated in the running head.
% If there are more than two authors, 'et al.' is used.
%
\institute{
Intel – Artificial Intelligence Products Group (AIPG)\and Technion – Israel Institute of Technology \\
\email{\href{mailto:yury.nahshan@intel.com}{yury.nahshan@intel.com};
\href{mailto:brian.chmiel@intel.com}{brian.chmiel@intel.com};
\href{mailto:chaimbaskin@cs.technion.ac.il}{chaimbaskin@cs.technion.ac.il};
\href{mailto:evgeniizh@campus.technion.ac.il}{evgeniizh@cs.technion.ac.il};
\href{mailto:ron.banner@intel.com}{ron.banner@intel.com};
\href{mailto:bron@cs.technion.ac.il}{bron@cs.technion.ac.il};
\href{mailto:avi.mendelson@cs.technion.ac.il}{avi.mendelson@cs.technion.ac.il} } }
%\end{comment}
%******************
\maketitle

\begin{abstract}
Neural network quantization enables the deployment of large models on resource-constrained devices. Current post-training quantization methods fall short in terms of accuracy for INT4 (or lower) but provide reasonable accuracy for INT8 (or above). In this work, we study the effect of quantization on the structure of the loss landscape. Additionally, we show that the structure is flat and separable for mild quantization, enabling straightforward post-training quantization methods to achieve good results. We show that with more aggressive quantization, the loss landscape becomes highly non-separable with steep curvature, making the selection of quantization parameters more challenging. Armed with this understanding, we design a method that quantizes the layer parameters jointly, enabling significant accuracy improvement over current post-training quantization methods.
Reference implementation is available at \href{https://github.com/ynahshan/nn-quantization-pytorch/tree/master/lapq}{the paper repository}.
\end{abstract}

\newcommand{\upperstar}[1]{\protect\accentset{*}{#1}}

%%%%%%%%% BODY TEXT

\input{sections/010_intro.tex}
\input{sections/020_related.tex}
\input{sections/030_loss_analysis.tex}

\input{sections/040_method.tex}

\input{sections/050_experiments.tex}

\input{sections/060_conclusion.tex}

%\subsubsection*{Author Contributions}

\subsubsection*{Acknowledgments}The research was funded by National Cyber Security Authority and the Hiroshi Fujiwara Technion Cyber Security Research Center.

\clearpage
% ---- Bibliography ----
%
% BibTeX users should specify bibliography style 'splncs04'.
% References will then be sorted and formatted in the correct style.
%
\bibliographystyle{splncs04}
\bibliography{lapq_eccv}

\newpage

\appendix{}

%%% FIGURE NUMBERING IN APPENDIX
\renewcommand\thefigure{\thesection.\arabic{figure}} 
\renewcommand\thetable{\thesection.\arabic{table}} 
\renewcommand\theequation{\thesection.\arabic{equation}}  
\setcounter{figure}{0}  
\setcounter{table}{0}

\crefalias{section}{appsec}
\crefalias{subsection}{appsec}
\crefalias{subsubsection}{appsec}

\input{sections/070_appendix.tex}
\end{document}

%% file: sections/010_intro.tex
\section{Introduction}
\label{sec:intro}

\begin{figure}[h!]
 \centering
    \begin{subfigure}{0.45\linewidth}
        \includegraphics[width=\linewidth]{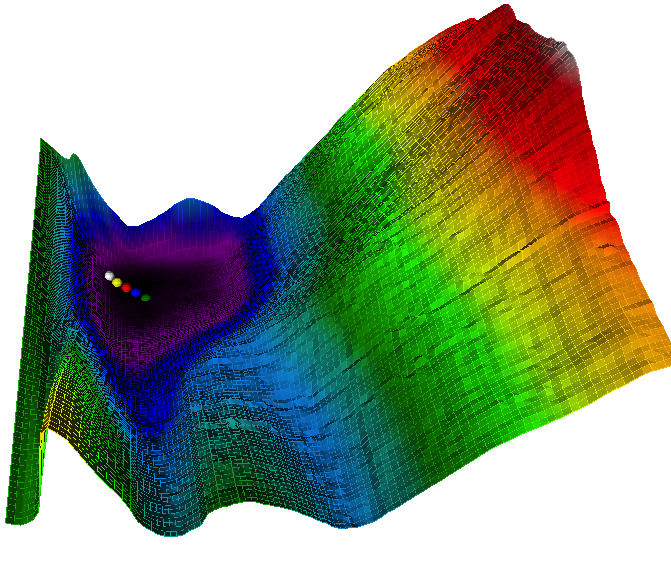}
        \caption{}
\label{fig:2bitMesh}
    \end{subfigure}
     \begin{subfigure}{0.45\linewidth}
        \includegraphics[width=\linewidth]{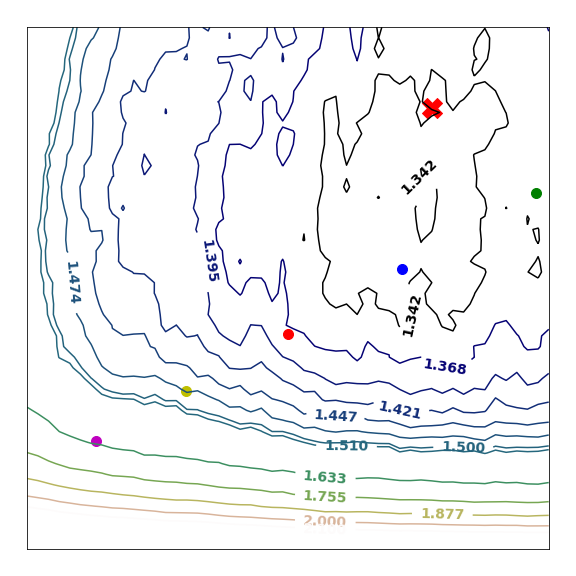}
         \caption{}
\label{fig:2bitContour}
    \end{subfigure}
\caption{A visualization of the loss surface: For every pair of quantization step sizes in the first two layers (x and y-axis), we estimate the cross-entropy loss  (z-axis) using one batch of 512 images (ResNet18 on ImageNet). The colored dots mark the quantization step sizes found by optimizing different p-norm metrics -- each layer is quantized in a way that minimizes the p-norm distance between the quantized tensor to its original full precision counterpart (e.g., MSE distortion). (a) Note the complex inseparable shape of the loss landscape. (b) A zoom-in of sub-figure (a) to the point where the location of the optimal cross-entropy loss (red cross) can be distinguished from the optimized p-norm solutions.}
% Cross-entropy loss on a calibration set (one batch of a training set of ImageNet) as a function of two clipping parameters, corresponding to the first two layers in ResNet-18, which are quantized to 2 bits. Colored dots mark solutions found by different local \AB{I don't understand what local means? Are we ever using any global optimization methods?} optimization methods. Note the complex inseparable shape of the loss landscape. (b) Area around the minimum (denoted a by red cross) of the same loss function of ResNet-18. \ez{it is the same, right?}}
\end{figure}

Deep neural networks (DNNs) are a powerful tool that have shown unmatched performance in various tasks in computer vision, natural language processing and optimal control, to mention only a few. The high computational resource requirements, however, constitute one of the main drawbacks of DNNs, hindering their massive adoption on edge devices. With the growing number of tasks performed on edge devices, e.g., smartphones or embedded systems, and the availability of dedicated custom hardware for DNN inference, the subject of DNN compression has gained popularity. 

One way to improve DNN computational efficiency  is to use lower-precision representation of the network, also known as quantization.
Most of the literature on neural network quantization involves training either 
from scratch \cite{bethge2019back,jin2019towards} or performing fine-tuning on a pre-trained full-precision model \cite{yang2019quantization,Hubara:2017:QNN:3122009.3242044}. 
While training is a powerful method to compensate for accuracy loss due to quantization, %\AM{often,}
it is often desirable to be able to quantize the model without training since this is both resource consuming and requires access to the data the model was trained on, which is not always available. 
%Consequently, it is desirable to apply quantization without fine-tuning the model or, at least, without fully training it from scratch. 
These methods are commonly referred to as \textit {post-training quantization} and usually require only a small calibration dataset. %Recently, post-training quantization has attracted much attention  because they do not require neither large amount of data nor complex training procedures. 
 Unfortunately, current  post-training methods are not very efficient, and most existing works only manage to quantize parameters to the 8-bit integer representation (INT8).

In the absence of a training set, these  methods typically aim at minimizing the local error introduced during the quantization process (e.g., round-off errors). Recently, a popular approach to minimizing this error has been to clip the tensor outliers. This means that peak values will incur a larger error, but, in total, this will reduce the distortion introduced by the limited resolution \cite{lee2018quantization,banner2018post,zhao2019improving}. Unfortunately, these schemes suffer from two fundamental drawbacks.

Firstly, it is hard or even impossible to choose an optimal metric for the network performance based on the tensor level quantization error. In particular, even for the same task, similar architectures may favor different objectives. Secondly, the noise in earlier layers might be amplified by successive layers, creating a dependency between quantizaton errors of different layers. This cross-layer dependency makes it necessary to jointly optimize the quantization parameters across all network layers; however, current methods optimize them separately for each layer. Figure 1 plots the challenging loss surface of this optimization process. 

Below, we outline the main contributions of the present work along with the organization of the remaining sections.  
\begin{description}
\item[$\bullet$] First we consider current layer-by-layer quantization methods where the quantization step size within each layer is optimized to accommodate the dynamic range of the tensor while keeping it small enough to minimize quantization noise. Although these methods optimize the quantization step size of each layer independently of the other layers, we observe strong interactions between the layers, explaining their suboptimal performance at the network level.

\item[$\bullet$] Accordingly, we consider network quantization as a multivariate optimization problem where the layer quantization step sizes are jointly optimized to minimize the neural network cross-entropy loss. We observe that layer-by-layer quantization identifies solutions in a small region around the optimum, where degradation is quadratic in the distance from it. We provide analytical justification as well as empirical evidence showing this effect. 

\item[$\bullet$] Finally, we propose to combine layer-by-layer quantization with multivariate quadratic optimization. Our method is shown to significantly outperform state-of-the-art methods on two different challenging tasks and six DNN architectures. 
\end{description}

The rest of the paper is organized as follows: \cref{sec:related} reviews the related work, \cref{sec:local} studies the properties of the loss function of the quantized network, \cref{sec:global} describes a proposed method, \cref{sec:exp} provides the experimental results, and \cref{sec:conclusion} concludes the paper.

%% file: sections/020_related.tex
\section{Related work}
\label{sec:related}

The approaches to neural network quantization can be divided roughly into two major categories  \cite{Krishnamoorthi2018whitepaper}: quantization-aware training, which introduces the quantization at some point during the training, and post-training quantization, where the network weights are not optimized during the quantization.
The recent progress in quantization-aware training has allowed the realization of results comparable to a baseline for as low as 2--4 bits per parameter \cite{baskin2018nice, zhang2018lq,  gong2019differentiable, yang2019quantization,jin2019towards} and show decent performance even for single bit (binary) parameters \cite{liu2019rbcn,peng2019bdnn,kim2020binaryduo}. 
The major drawback of quantization-aware methods is the necessity for a vast amount of labeled data and high computational power. 
%It poses an obstacle to using them in embedded platforms, where both terms unfeasible due to power limitations. 
Consequently, post-training quantization is widely used in existing embedded hardware solutions. Most work that proposes hardware-friendly post-training schemes has only managed to get to 8-bit quantization without significant degradation in performance or requiring substantial modifications in exisiting hardware.

One of the first post-training schemes, proposed by \citet{gong2018highly}, used min-max ($\ell_\infty$ norm) as a threshold for 8-bit quantization. It produces a small performance degradation and can be deployed efficiently on the hardware. Afterward, better schemes involving the choosing of the clipping value,  such as minimizing the Kullback-Leibler divergence  \cite{migacz20178}, assuming the known distribution of weights \cite{banner2018post}, or 
 minimizing quantization MSE iteratively \cite{choukroun2019low},  were proposed.

%\paragraph{Changing distribution}
Another efficient approach quantization is to change the distribution of the tensors, making it more suitable for quantization, such as  equalizing the weight ranges
\cite{nagel2019data} or splitting outlier channels
\cite{zhao2019improving}.

%\paragraph{Bias correction}
Recent works noted that the quantization process introduces a bias into distributions of the parameters and focused on correction of this bias.  \citet{finkelstein2019fighting} addressed a problem of MobileNet quantization. They claimed that the source of degradation was shifting in the mean activation value caused by inherent bias in the quantization process and proposed a scheme for fixing this bias. Multiple alternative schemes for the correction of quantization bias were proposed, and those techniques are widely applied in state-of-the-art quantization approaches
\cite{banner2018post,finkelstein2019fighting,nagel2019data,fang2020nearlossless}. In our work, we utilize the bias correction method by developed \citet{banner2018post}. 

%\paragraph{Fine-grained quantization}
While the simplest form of quantization applies a single quantizer to the whole tensor (weights or activations), finer quantization allows reduction of performance degradation. Even though this approach usually boosts performance significantly \cite{Mellempudi2017ternary,banner2018post,faraone2018syq,choukroun2019low,lee2018quantization}, it requires more parameters and special hardware support, which makes it unfavorable for real-life deployment. 

%\paragraph{Clustering and non-uniform quantization}
One more source of performance improvement is using more sophisticated ways to map values to a particular bin. This includes clustering of the tensor entry values 
\cite{nayak2019bit}  and non-uniform quantization
\cite{fang2020nearlossless}. By allowing a more general quantizer, these approaches provide better performance than uniform quantization, but also require hardware support for efficient inference, and thus, similarly to fine-grained quantization, are less suitable for deployment on consumer-grade hardware.

%\paragraph{Separability}
To the best of our knowledge, previous works did not take into account the fact that the loss might be not separable during optimization, usually performed per layer or even per channel. Notable exceptions are \citet{gong2018highly} and \citet{zhao2019improving}, who did not show any kind of optimization.
\citet{nagel2019data} partially addressed the lack of separability by treating pairs of consecutive layers together.

%% file: sections/030_loss_analysis.tex
\section{Loss landscape of quantized DNNs}
\label{sec:local}
In this section, we introduce the notion of separability of the loss function. We study the separability and the curvature of the loss function and show how quantization of DNNs affect these properties. Finally, we show that during aggressive quantization, the loss function becomes highly non-separable with steep curvature, which is unfavorable for existing post-training quantization methods. Our method addresses these properties and makes post-training quantization possible at low bit quantization.

We focus on symmetric uniform quantization with a fixed number of bits $M$ for all layers and quantization step size $\Delta$ that maps a continuous value $x\in \mathbb{R}$ into a discrete representation, %$Q_{\Delta}(x)=i\cdot \Delta$ with $ i\in \{0,1,..., 2^M -1\}$.  The index $i$ is expressed as follows: 
\begin{align}
Q_{\Delta, M}(x) = \begin{cases}
-2^{M-1} \Delta & x < -2^{M-1} \Delta\\
\left\lfloor\dfrac{x}{\Delta} \right\rceil \Delta  &  \abs{x} \leqslant 2^{M+1}\Delta\\
+2^{M-1} \Delta & x > + 2^{M-1} \Delta.
\end{cases}
\end{align}

By constraining the range  of  $x$ to $[-c, c]$, the connection between $c$ and $\Delta$ is given by: 
\begin{align}
    \Delta = \frac{2 c}{2^{M-1}}
\end{align}

In the case of activations, we limit ourselves to the ReLU function, which allows us to choose a quantization range of $[0, c]$. In such cases, the quantization step $\Delta$ is given by:
\begin{align}
    \Delta = \frac{c}{2^{M-1}}
\end{align}

% \YN{In the separability section instead of additive noise model we define multiplicative noise. This miss leading, consider remove this part (yellow).}

% \textcolor{GreenYellow}{

% \RB{what's that? I didnt read the paper. what is $\phi$? the additive noise model is well established for sufficient resolution...}
% This assumption is legitimate, since the sufficient and necessary condition for quantization error to be white and uniform is vanishing characteristic function of the input \cite{sripad1977necessary}:
% \BCH{not clear enough}
% \begin{align}
%     \forall n\neq 0 \quad \phi_x\qty(\frac{2\pi n}{\Delta}) = \mathbb{E}_x\qty[\exp(\frac{2\pi i nx}{\Delta})] = 0.
% \end{align}
% For large amount of quantization bins, $\nicefrac{x}{\Delta}$ is close to integer, and thus the condition is approximately satisfied, which was also confirmed empirically for NN feature maps \cite{baskin2018nice}.
% }

% \begin{figure}
% \centering
%     \begin{subfigure}{0.49\linewidth}
%         \includegraphics[width=\linewidth]{figures/quant_matmul_sep_2bit.pdf}
%         \caption{2 bit}
%     \end{subfigure}
%     \begin{subfigure}{0.49\linewidth}
%         \includegraphics[width=\linewidth]{figures/quant_matmul_sep_4bit.pdf}
%         \caption{4 bit}
%     \end{subfigure}
%  \caption{}
% \label{fig:quant_matmul_mse}
% \end{figure}

\begin{figure*}
 \centering
    \begin{subfigure}{0.28\linewidth}
        \includegraphics[width=\linewidth]{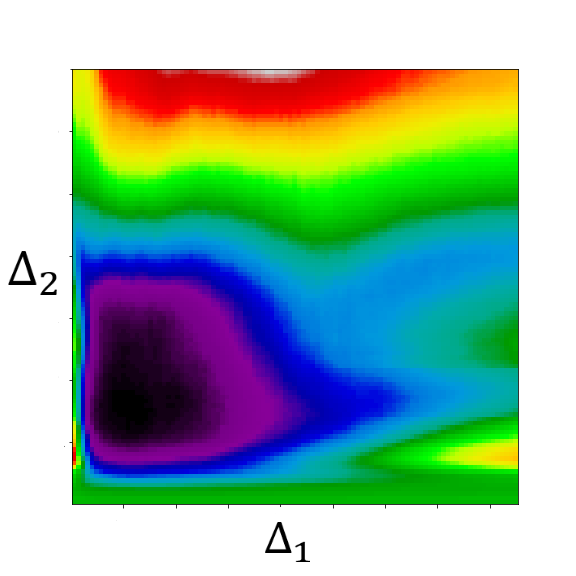}
        \subcaption{2 bit}
        \label{heatmap_w2bit}
    \end{subfigure}
    \begin{subfigure}{0.28\linewidth}
        \includegraphics[width=\linewidth]{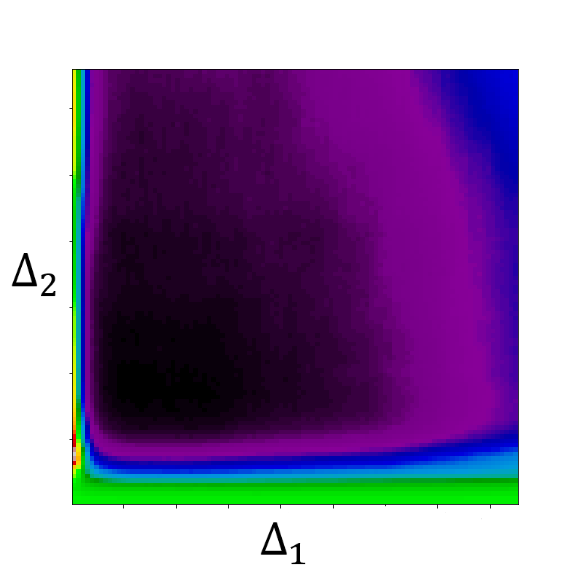}
        \subcaption{3 bit}
        \label{heatmap_w3bit}
    \end{subfigure}
    \begin{subfigure}{0.28\linewidth}
        \includegraphics[width=\linewidth]{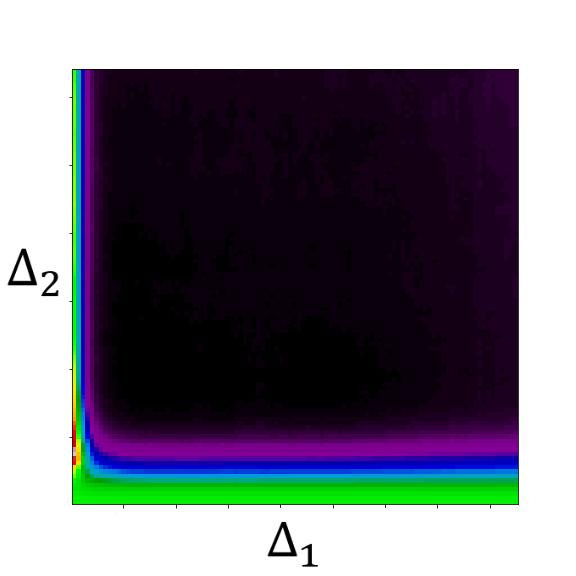}
        \subcaption{4 bit}
        \label{heatmap_w4bit}
    \end{subfigure}
        \begin{subfigure}{0.09\linewidth}
        \vspace{-0.6cm}
        \includegraphics[width=\linewidth, height=90pt]{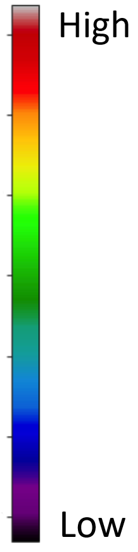}
    \end{subfigure}
 \caption{A visualization of the loss surface showing  complex interactions between quantization step sizes of the first two layers in ResNet18. At higher bitwidth, the quantization step sizes $\Delta_1$ and $\Delta_2$ are smaller, thus introducing a smaller level of noise. \cref{Interaction} suggests a weak interaction between the first and second layers in that case, which becomes evident at 4-bit precision (the effect on the loss due to the quantization of the first layer with $\Delta_1$ is almost independent of the effect that is achieved with $\Delta_2$). On the other hand, at lower bitwidths (i.e., higher quantization noise), we start seeing interactions between the quantization processes of the two layers. }
\label{fig:heatmap}
\end{figure*}

\subsection{Separable optimization}
Suppose the loss function of the network $\mathcal{L}$ depends on
a certain set of variables (weights, activations, etc.), which we denote by a vector $\mathbf{v}$. We would like to measure the effect of adding quantization noise to this set of vectors. In the following we show that for sufficiently small quantization noise, we can treat it as an additive noise vector $\bm{\varepsilon}$, allowing coordinate-wise optimization. However, when quantization noise is increased, the degradation in one layer is associated with other layers, calling for more laborious non-separable optimization techniques. From the Taylor expansion: 
\begin{equation}
\Delta \mathcal{L} = 
\mathcal{L}\qty(\mathbf{v}+\bm{\varepsilon})-\mathcal{L}\qty(\mathbf{v})=
\pdv{\mathcal{L}}{\mathbf{v}}^{\top}\bm{\varepsilon}+
\bm{\varepsilon}^{\top}\pdv[2]{\mathcal{L}}{\mathbf{v}}\bm{\varepsilon}+\order{\norm{\bm{\varepsilon}}^3}
\label{taylor}
\end{equation}
When the quantization error $\bm{\varepsilon}$ is sufficiently small, higher-order terms can be neglected so that degradation   $\Delta \mathcal{L}$ can be approximated as a sum of the quadratic functions, 
\begin{equation}
\Delta \mathcal{L}  =\pdv{\mathcal{L}}{\mathbf{v}}^{\top}\bm{\varepsilon}+\bm{\varepsilon}^{\top}\pdv[2]{\mathcal{L}}{\mathbf{v}}\bm{\varepsilon}
=  \sum_i^n \pdv{\mathcal{L}}{{v_i}}\cdot \varepsilon_i + \sum_i^n \sum_j^n \pdv{\mathcal{L}}{v_i}{v_j}\varepsilon_i \cdot \varepsilon_j.
\label{taylor_quadratic}
\end{equation}
One can see from \cref{taylor_quadratic} that when quantization error $\norm{\bm{\varepsilon}} ^{2}$ is sufficiently small, the overall degradation $\Delta \mathcal{L}$ can be approximated as a sum of $N$ independent separable degradation processes as follows:  
\begin{equation}
\Delta \mathcal{L}  
\approx  \sum_i^n \frac{\partial\mathcal{L}}{\partial{v_i}}\cdot \varepsilon_i 
\label{taylor_linear}
\end{equation}
On the other hand, when $\norm{\bm{\varepsilon}} ^{2}$ is larger, one needs to take into account the interactions between different layers, corresponding to the second term in \cref{taylor_quadratic} as follows:  

\begin{equation}
 \text{QIT}=\sum_i^n \sum_j^n \pdv{\mathcal{L}}{v_i}{v_j}\varepsilon_i \cdot \varepsilon_j,
\label{Interaction}
\end{equation}
where QIT refers to the quantization interaction term.

In \cref{fig:heatmap} we provide a visualization of these interactions at 2, 3 and 4 bitwidth representations. 

\begin{figure}
\centering
\includegraphics[width=0.5\linewidth]{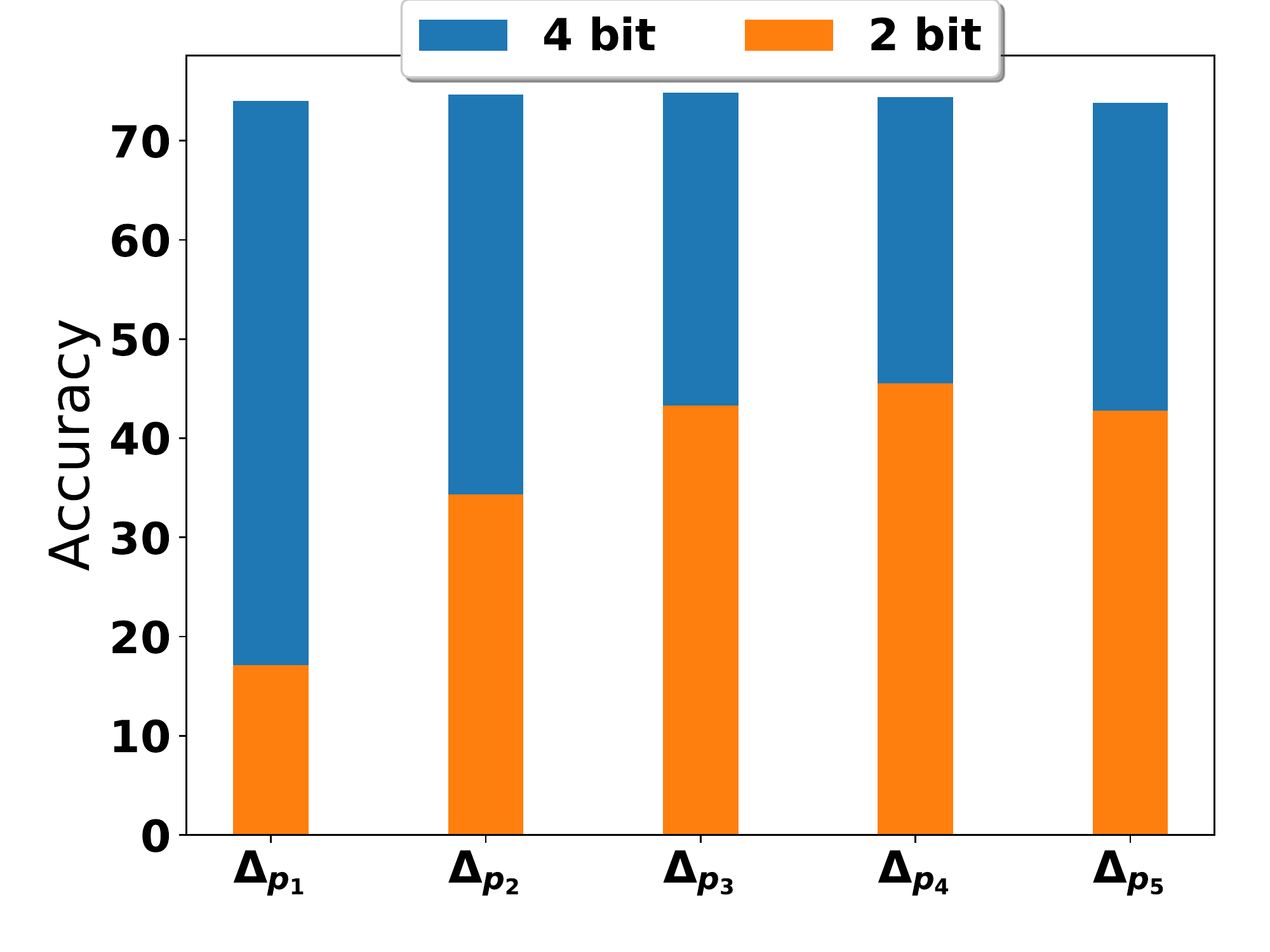}
 \caption{Accuracy of ResNet-50 quantized to 2 and 4 bits, respectively. Quantization steps are chosen such that they minimize the $L_p$ norm of the quantization error for different values of $p$.}
\label{fig:pNorms_p_vs_acc}
\end{figure}

\subsection{Curvature}
We now analyze how the steepness of the curvature of the loss function with respect to the quantization step changes as the quantization error increases. We will show that at aggressive quantization, the curvature of the loss becomes steep, which is unfavorable for the methods that aim to minimize quantization error on the tensor level.

We start by defining a measure to quantify curvature of the loss function with respect to quantization step size $\Delta$. Given a quantized neural network, we denote $\mathcal{L}(\Delta_1, \Delta_2, \dots, \Delta_n)$ the loss with respect to quantization step size $\Delta_i$ of each individual layer. Since $\mathcal{L}$ is twice differentiable with respect to $\Delta_i$ we can calculate the Hessian matrix:
\begin{align}
    H[\mathcal{L}]_{ij} = \pdv{\mathcal{L}}{\Delta_i}{\Delta_j}.
\end{align}
To quantify the curvature, we use Gaussian curvature \cite{goldman2005curvature}, which is given by:
\begin{align}
    K[\mathcal{L}](\Delta) = \frac{\det\qty(H[\mathcal{L}](\Delta))}{\qty( \norm{\nabla \mathcal{L}(\Delta)}_2^2 + 1)^2},
\end{align}
where  $\Delta=\qty(\Delta_1, \Delta_2, \dots, \Delta_n)$.
% At minimum $\grad f(x)=0$ and thus
% \begin{align}
%     K[f](x) = \det\qty(H[f](\vb{x}))
% \end{align}
We calculated the Gaussian curvature at the point that minimizes the $L_2$ norm of the quantization error and acquired the following values:
\begin{align}
    K\qty[\mathcal{L}_{\text{4 bit}}](\Delta) &=6.7 \cdot 10^{-25}\\
    K\qty[\mathcal{L}_{\text{2 bit}}](\Delta) &=0.58,
\end{align}
This means that the flat surface for 4 bits, shown in \cref{fig:heatmap}, is a generic property of the fine-grained quantization loss and not of the specific layer choice. Similarly, we conclude that coarser quantization generally has steeper curvature than more fine-grained quantization.

Moreover, the Hessian matrix provides additional information regarding the coupling between different layers. As could be expected, adjacent off-diagonal terms have higher values than distant elements, corresponding to higher dependencies between clipping parameters of adjacent layers (the Hessian matrix is presented in \cref{fig:hessian} in the Appendix).

\begin{figure}
    \centering
    \includegraphics[width=0.5\linewidth]{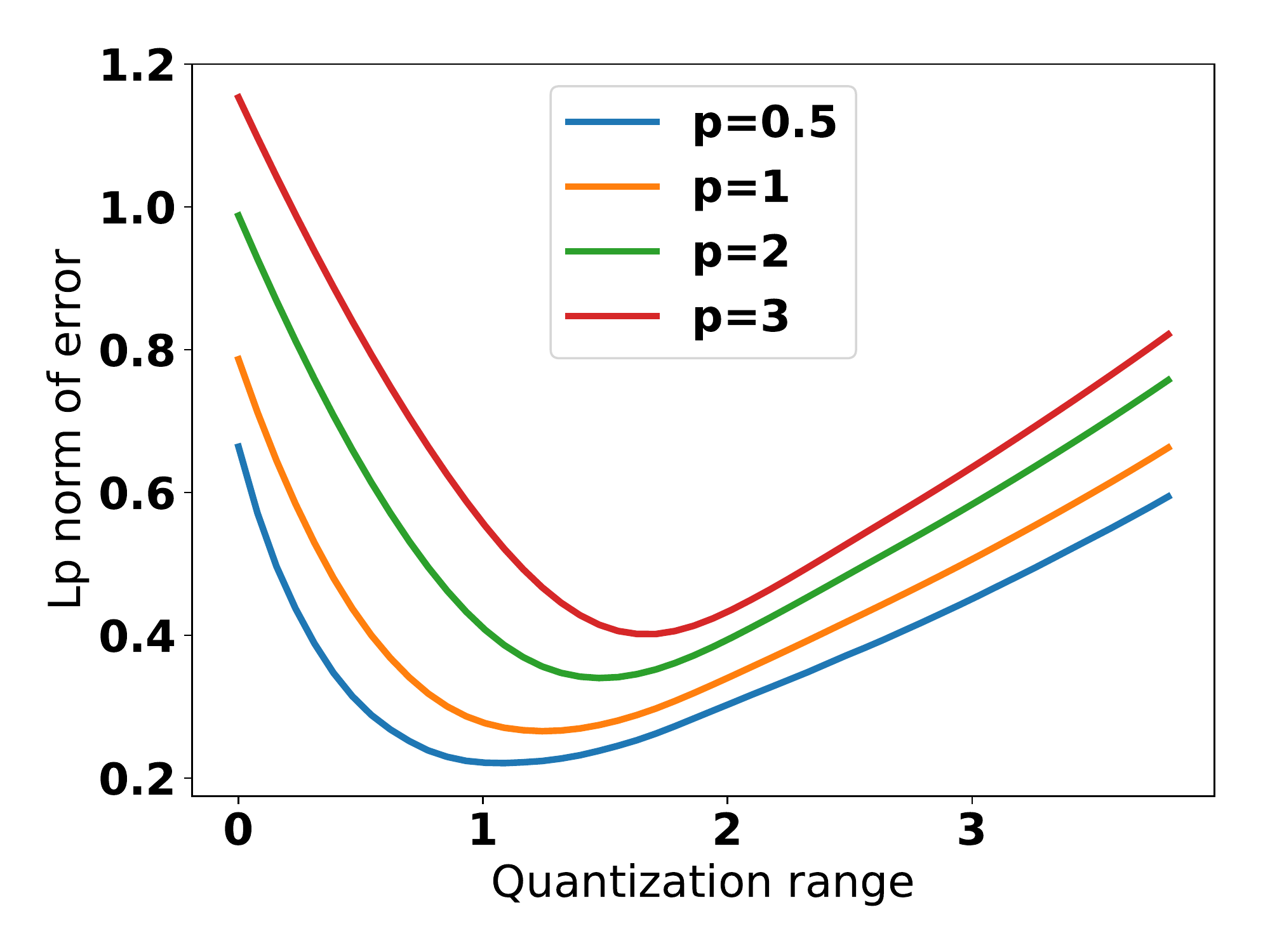}
    \caption{The $L_p$ norm of the quantization error generated by optimizing for different values of $p$ according to \cref{eq:pnorm}. For different $L_p$ metrics, the optimal quantization step is different.}
    \label{fig:p_vs_lpnorm}
\end{figure}

When the curvature is steep, even small changes in quantization step size may change the results drastically. To justify this hypothesis experimentally, in \cref{fig:pNorms_p_vs_acc}, we evaluate the accuracy of ResNet-50 for five different quantization steps. We choose quantization steps which minimize the $L_p$ norm of the quantization error for different values of $p$.

While at 4-bit quantization, the accuracy is almost not affected by small changes in the quantization step size, at 2-bit quantization, the same changes shift the accuracy by more than 20\%. Moreover, the best accuracy  is obtained with a quantization step that minimizes $L_{3.5}$ and not the MSE, which corresponds to $L_2$ norm minimization.

%% file: sections/040_method.tex
\section {Loss Aware Post-training Quantization  (LAPQ)}
\label{sec:global}
In the previous section, we showed that the loss function $\mathcal{L}(\Delta)$, with respect to the quantization step size $\Delta$, has a complex, non-separable landscape that is hard to optimize.
In this section, we suggest a method to overcome this intrinsic difficulty. Our optimization process involves three consecutive steps.

In the first phase, we find the quantization step $\Delta_p$ that minimizes the $L_p$ norm of the quantization error of the individual layers for several different values of $p$. Then, we perform quadratic interpolation to approximate an optimum of the loss with respect to $p$. Finally, we jointly optimize the parameters of all layers acquired on the previous step by applying a gradient-free optimization method \cite{1964Powell}. The pseudo-code of the whole algorithm is presented in \cref{alg:powel}. \cref{fig:lapq_intuition} provides the algorithm visualization.

\begin{figure}
\centering
    \begin{subfigure}{0.49\linewidth}
        \includegraphics[width=\linewidth]{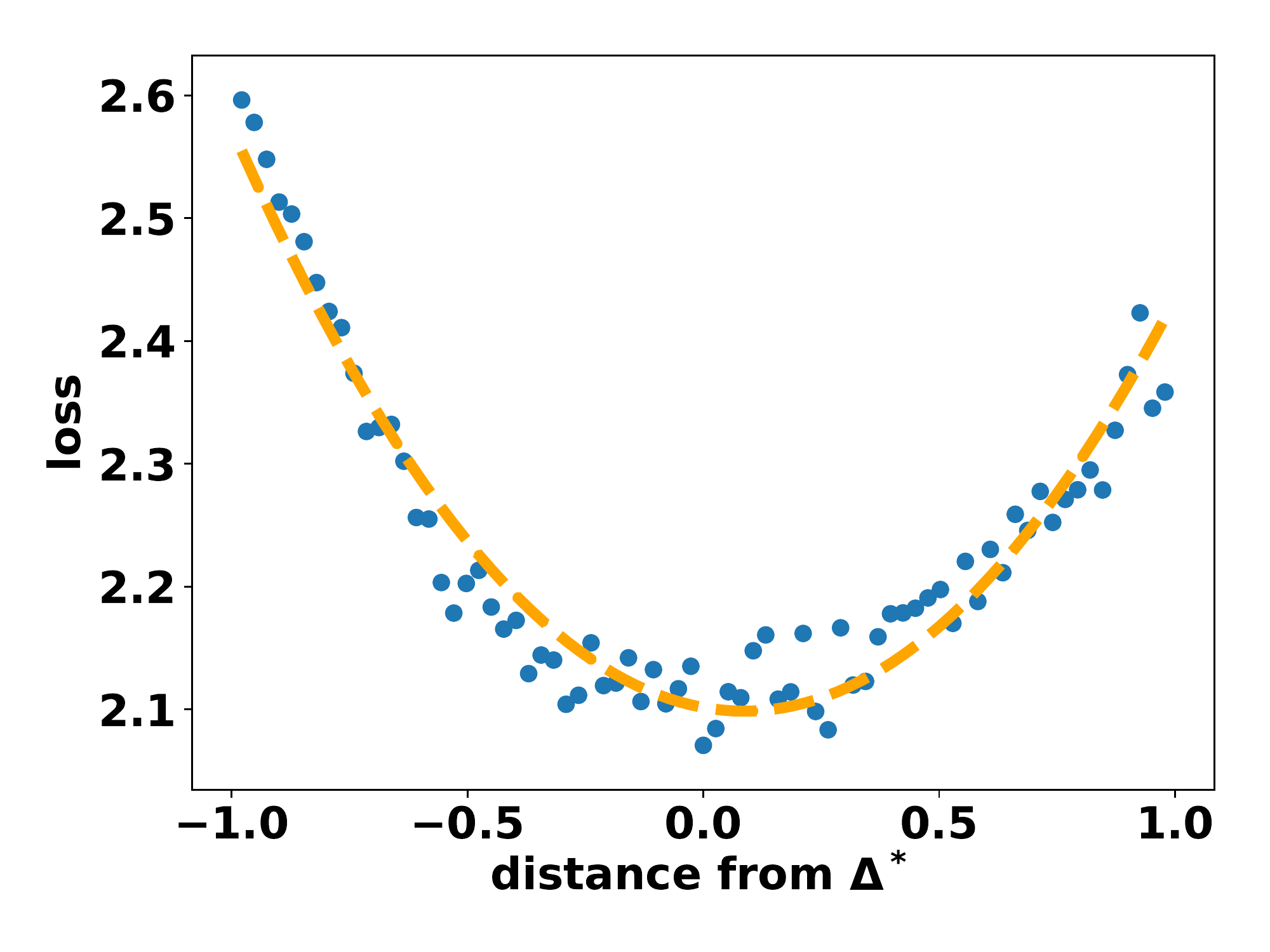}
    \end{subfigure}
    \begin{subfigure}{0.49\linewidth}
        \includegraphics[width=\linewidth]{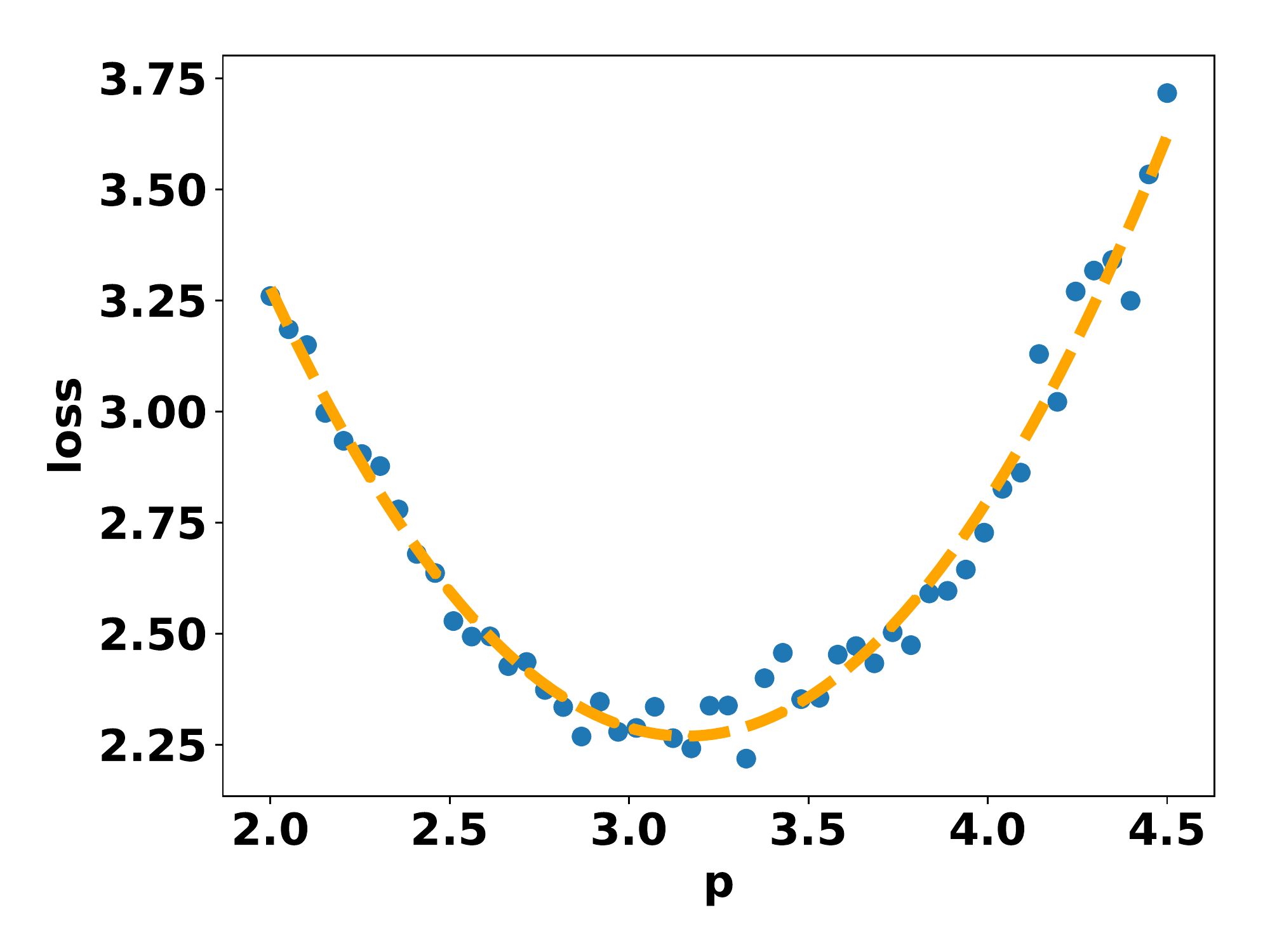}
    \end{subfigure}
 \caption{Loss of ResNet-18 quantized with different quantization steps. The orange dotted line shows quadratic interpolation: (a) on the left with respect to the distance from optimal quantization step $\Delta^*$ and (b) on the trajectory defined by $L_p$ norm minimization.}
\label{fig:pNorms_res18_res50}
\end{figure}

\subsection {Layer-wise optimization} 
Our method starts by minimizing the $L_p$ norm of the quantization error of weights and activations in each layer with respect to clipping values:
\begin{align}
    e_p(\Delta) = \qty\bigg(\norm{Q_{\Delta}(X) - X}^p )^{\nicefrac{1}{p} }.
\label{eq:pnorm}
\end{align}

Given a real number $p>0$, the set of optimal quantization steps $\Delta_p=\{\Delta_{1_p},\Delta_{2_p},\dots,\Delta_{n_p}\}$, according to \cref{eq:pnorm}, minimizes the quantization error within each layer. In addition, optimizing the quantization error allows us to get $\Delta_p$ in the vicinity of the optimum $\Delta^*$ . Different values of $p$ result in different quantization step sizes $\Delta_p$, which are still optimal under some metric $L_p$ due to the trade-off between clipping and quantization error (\cref{fig:p_vs_lpnorm}).

\subsection{Quadratic approximation}
Assuming a quantization step size $\Delta$ in the vicinity of the optimal quantization step $\Delta^*$, the loss function can be approximated with a Taylor series as follows:
\begin{align}
   &\mathcal{L}(\Delta) - \mathcal{L}(\Delta^*) =\\
   =&(\Delta^* - \Delta )^\top \nabla\mathcal{L}(\Delta^*) + \frac{1}{2} (\Delta^* - \Delta)^\top\mathbf{H}(\Delta^*)(\Delta^* - \Delta ) + \order{\norm{\Delta^* - \Delta }^3},
\label{eq:taylor_loss_full}
\end{align}
where $\mathbf{H}(\Delta^*)$ is the Hessian matrix with respect to $\Delta$.
Since $\Delta^*$ is a minimum, the first derivative vanishes and we acquire a quadratic approximation of $\mathcal{L}$,
\begin{equation}
   \Delta \mathcal{L}  = \mathcal{L}(\Delta) - \mathcal{L}(\Delta^*) \approx \frac{1}{2} (\Delta^* - \Delta )^\top\mathbf{H}(\Delta^*)(\Delta^* - \Delta) 
\label{eq:taylor_loss_full1}
\end{equation}
%Equation \ref{eq:taylor_loss_full1} shows that the degradation $\Delta \mathcal{L}(C)$ is quadratic in the distance from $\accentset{*}{C}$. 

\begin{figure}
    \centering
    \includegraphics[width=0.9\linewidth]{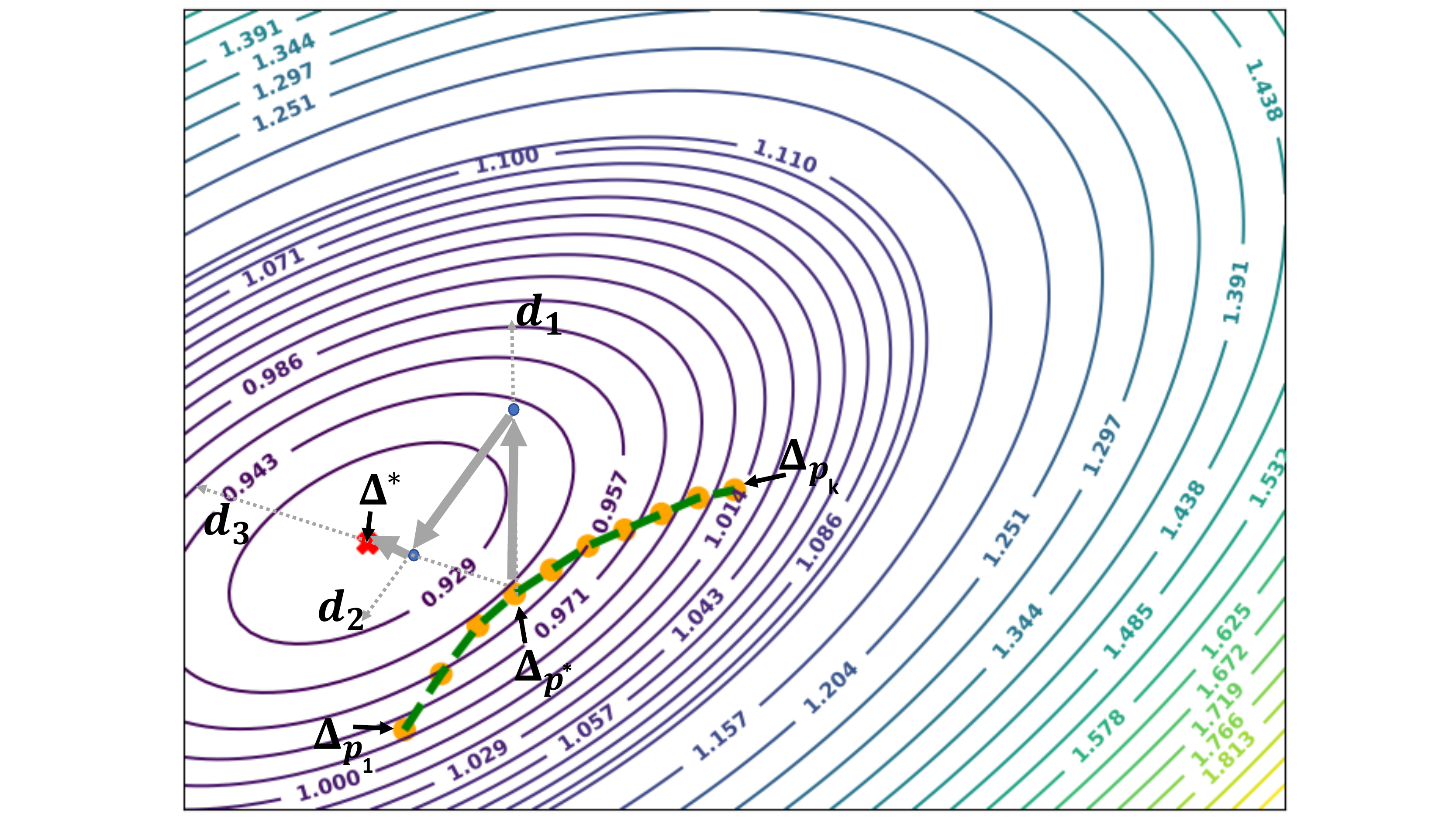}
    \caption{Illustration of the LAPQ algorithm. Yellow dots correspond to the quantization step size $\{\Delta_p\}$, which minimizes the $L_p$ norm of the quantization error. $\{\Delta_p\}$ build a trajectory in the vicinity of the minimum $\Delta^*$. The optimal quantization step size on that trajectory $\Delta_{p^*}$ is used as a starting point for a joint optimization algorithm (Powell's). Vectors $\{d_1, d_2, d_3\}$ demonstrate the first iteration of Powell's method that brings us closer to the global minimum $\Delta^*$. Better viewed in color.}
    \label{fig:lapq_intuition}
\end{figure}

Our method exploits this quadratic property for optimization. \cref{fig:pNorms_res18_res50}(a) demonstrates empirical evidence of such a quadratic relationship for ResNet-18 around the optimal quantization step $\Delta^*$ obtained by our method. First, we sample a few data points $\{\Delta_p\}$ to build a trajectory on the graph of $\mathcal{L}(\Delta)$ (orange points in \cref{fig:lapq_intuition}). Then, we use the prior quadratic assumption to approximate the minimum of the $\mathcal{L}$ on that trajectory by fitting a quadratic function $f(p)$ to the sampled $\Delta_p$. Finally, we minimize $f(p)$ and use the optimal quantization step size $\Delta_{p^*}$ as a starting point for a gradient-free joint optimization algorithm, such as Powell's method \cite{1964Powell}, to minimize the loss and find $\Delta^*$.
%\textcolor{red}{This clipping values could be found using SGD, but this requires specific software and hardware supporting training in higher precision, e.g., can't be deployed on dedicated inference-only accelerators. }\YN{Are you sure this is the right place for this quote? IMHO it more suitable for introduction. Mentioning SGD after we present our method feels like shooting yourselves in the leg.}

\subsection{Joint optimization}

% \begin{figure}
% \centering
%         \begin{subfigure}{0.49\linewidth}
%             \includegraphics[width=\linewidth]{figures/powell_explanation1.pdf}
%             \caption{}
%             \label{fig:powellExplain1}

%         \end{subfigure}
%         \begin{subfigure}{0.49\linewidth}
%             \includegraphics[width=\linewidth]{figures/powell_explanation2.pdf}
%             \caption{}
%              \label{fig:powellExplain2}
%         \end{subfigure}
%  \caption{Intuition of Powell's method. \textbf{(a)} First iteration: Starting point is $C_0$ and the initial search direction vectors $\{d_1,d_2\}$. The optimization is in the conjugate direction ($d_3$).
%  \textbf{(b)} Second iteration,  the previous conjugate direction $d_3$ was added as part of the directions search. Optimization is in the conjugate direction $d_4$ which is the direction that minimize the function. \ez{consider removing}\BCH{Disagree}}
% \label{fig:powellExplain}
% \end{figure}

By minimizing both the quantization error and the loss using quadratic interpolation, we get a decent approximation of the global minimum $\Delta^*$. Due to steep curvature of the minimum, however, for a low bitwidth quantization, even a small error in the value of $\Delta$ leads to performance degradation. Thus, we use a gradient-free joint optimization, specifically an iterative algorithm (Powell's method \cite{1964Powell}), to further optimize $\Delta_{p^*}$.

\begin{algorithm}[h!]
	\caption{LAPQ}
	\label{alg:powel}
	\begin{algorithmic}[1]
		\State \underline{\textbf{Layer-wise optimization: }}
		\For{$p = \{p_1, p_2, \dots p_k\}$}
    			 \State $ \vb{C}_{p} \gets \arg\min\limits_{C} \qty(\norm{Q_{\Delta,c}(x) - x}^p)^{\nicefrac{1}{p}}$ for every layer
			
        	\EndFor

        \State \underline{\textbf{Quadratic approximation: }}
                \State $\Delta_{p^*} \gets \arg\min\limits_{\Delta_{p}}  \mathcal{L}(\Delta_{p})$  using quadratic interpolation of $\Delta_{p}$

		\State \underline{\textbf{Joint optimization (Powell's):} }
		\State Starting point $t_0 \gets  \Delta_{p^*}$. 
	%	\State Optimize with Powell's method \cite{1964Powell}
        \State Initial direction vector $D = \qty{d_1, d_2, \dots , d_N}$.
       % \State \underline{\textbf{Multi-variable optimization - Powell's method \cite{1964Powell}:} }
        \While{not converged}
    		\For{$k=1\dots N$}
        		\State $\lambda_k \gets \arg\min\limits_\lambda \mathcal{L}(t_{k-1} + \lambda_k d_k)$
        		\State $t_k \gets t_{k-1} + \lambda_k d_k$\
        	\EndFor
    		\For{$j=1\dots N-1$}
        		\State $d_j \gets d_{j+1}$
        	\EndFor	
        	\State $d_N \gets t_N - t_0$
    		\State $\lambda_N \gets \arg\min\limits_\lambda \mathcal{L}(t_{N} + \lambda_N d_N)$
    		\State $t_0 \gets t_{0} + \lambda_N d_N$\

    	\EndWhile
    	\State $\Delta^* \gets t_0$
		\State \textbf{return} $\Delta^*$ 
	\end{algorithmic}
\end{algorithm}

At every iteration, we optimize the set of parameters, initialized by $\Delta_{p^*}$. Given a set of linear search directions $D = \qty{d_1, d_2, \dots , d_N}$, the new position $\Delta_{t+1}$ is expressed by the linear combination of the search directions as following $\Delta_t + \sum_i{\lambda_id_i}$. The new displacement vector $\sum_i{\lambda_id_i}$ becomes part of the search directions set, and the search vector, which contributed most to the new direction, is deleted from the search directions set. For further details, see \cref{alg:powel}.

% In \cref{fig:powellExplain} we show 2 iterations of Powell's method for a 2 dimensional example. In the first iteration, \cref{fig:powellExplain1}, the search is across 2 initial search directions $\{d_1.d_2\}$. The new position $C_3$ is a linear combination of these directions and $d_3$ is added to the search directions vector. In the second interation, \cref{fig:powellExplain2}, the optimal direction ,$d_4$, is found which is a linear combination of $d_2$ and $d_3$.

% \paragraph{Bias correction} 
%

%% file: sections/050_experiments.tex
\section{Experimental Results}
\label{sec:exp}

In this section we conduct extensive evaluations and offer a comparison to prior art of the proposed method on two challenging benchmarks, image classification on ImageNet  and a recommendation system on NCF-1B. 
In addition, we examine the impact of each part of the proposed method on the final accuracy.
In all experiments we first calibrate the optimal clipping values on a small held-back calibration set using our method  and then evaluate the validation set.

\subsection{ImageNet}

\begin{table}
\caption{Comparison with other methods. {\color[HTML]{999903} MMSE} refers to minimization of mean square error.}% Note that ACIQ is not a search based method, which allows the clipping thresholds to be adjusted per channel. Note that our method performs layer-wise quantization, comparison to methods that require special hardware support (e.g., channel/kernel-wise quantization) is out of scope of this work.}
\label{tab:model_comparison}
\begin{subtable}{0.49\textwidth}
\scalebox{0.82}{
\begin{tabular} {lclc}
\toprule
\textbf{Model} & \textbf{W/A} & \textbf{Method} & \textbf{Accuracy(\%)} \\ \midrule
 & 32 / 32 & FP32 & 69.7 \\ \cline{2-4} 
 &  & {\color[HTML]{0475C1} LAPQ} (Ours) & \textbf{68.8} \\ 
 &  & {\color[HTML]{FF7C37} DUAL}  \cite{choukroun2019low} & 68.38 \\
 &  &{\color[HTML]{3F7E31} ACIQ}  \cite{banner2018post} & 65.528 \\ 
 & \multirow{-4}{*}{8 / 4} & {\color[HTML]{999903} MMSE} & 68 \\ \cline{2-4} 
 &  & {\color[HTML]{0475C1} LAPQ} (Ours) & \textbf{66.3} \\ 
 & & {\color[HTML]{3F7E31} ACIQ}  \cite{banner2018post} & 52.476 \\ 
& \multirow{-3}{*}{8 / 3} & {\color[HTML]{999903} MMSE} & 63.3 \\ \cline{2-4} 

 &  & {\color[HTML]{0475C1} LAPQ} (Ours) & \textbf{60.3} \\ 
  &  & {\color[HTML]{3F7E31} ACIQ}  \cite{banner2018post} & 4.1 \\ 

 &  & {\color[HTML]{9A0000} KLD}  \cite{migacz20178} & 31.937 \\ 
\multirow{-12}{*}{ResNet-18} & \multirow{-4}{*}{4 / 4} & {\color[HTML]{999903} MMSE} & 43.6 \\ \midrule
 & 32 / 32 & FP32 & 76.1 \\ \cline{2-4} 
 &  & {\color[HTML]{0475C1} LAPQ} (Ours) & \textbf{74.8} \\ 
 &  & {\color[HTML]{FF7C37} DUAL}  \cite{choukroun2019low} & 73.25 \\ 

 &  & {\color[HTML]{3F7E31} ACIQ}  \cite{banner2018post} & 68.92 \\ 
  & \multirow{-4}{*}{8 / 4} & {\color[HTML]{999903} MMSE} & 74 \\ \cline{2-4} 
 &  & {\color[HTML]{0475C1} LAPQ} (Ours) & \textbf{70.8} \\ 
& & {\color[HTML]{3F7E31} ACIQ}  \cite{banner2018post} & 51.858 \\ 
 & \multirow{-3}{*}{8 / 3} & {\color[HTML]{999903} MMSE} & 66.3 \\ \cline{2-4} 
 &  & {\color[HTML]{0475C1} LAPQ} (Ours) & \textbf{70} \\ 
   &  & {\color[HTML]{3F7E31} ACIQ}  \cite{banner2018post} & 3.3 \\ 

 &  & {\color[HTML]{9A0000} KLD}  \cite{migacz20178} & 46.19 \\ 
\multirow{-12}{*}{ResNet-50} & \multirow{-4}{*}{4 / 4} & {\color[HTML]{999903} MMSE} & 36.4 \\  \bottomrule
\end{tabular}
}
\end{subtable}%
\hfill
\begin{subtable}[h]{0.49\textwidth}
\scalebox{0.82}{

\begin{tabular}{lclc}
\toprule
\textbf{Model} & \textbf{W/A} & \textbf{Method} & \textbf{Accuracy(\%)} \\ \midrule
 & 32 / 32 & FP32 & \multicolumn{1}{c}{77.3} \\ \cline{2-4} 
 &  & {\color[HTML]{0475C1} LAPQ} (Ours) & 73.6 \\ 
 &  & {\color[HTML]{FF7C37} DUAL}  \cite{choukroun2019low} & \textbf{74.26} \\ 
 & & {\color[HTML]{3F7E31} ACIQ}  \cite{banner2018post} & 66.966 \\ 
&  \multirow{-4}{*}{8 / 4} & {\color[HTML]{999903} MMSE} & 72 \\\cline{2-4} 
 &  & {\color[HTML]{0475C1} LAPQ} (Ours) & \textbf{65.7} \\ 
 & & {\color[HTML]{3F7E31} ACIQ}  \cite{banner2018post} & 41.46 \\ &  \multirow{-3}{*}{8 / 3}  & {\color[HTML]{999903} MMSE} & 56.7 \\
 \cline{2-4} 
 &  & {\color[HTML]{0475C1} LAPQ} (Ours) & \textbf{59.2} \\ 
   &  & {\color[HTML]{3F7E31} ACIQ}  \cite{banner2018post} & 1.5 \\ 

 &  & {\color[HTML]{9A0000} KLD}  \cite{migacz20178} & 49.948 \\ 
\multirow{-12}{*}{ResNet-101} & \multirow{-4}{*}{4 / 4} & {\color[HTML]{999903} MMSE} & 9.8 \\ \midrule
 & 32 / 32 & FP32 & \multicolumn{1}{c}{77.2} \\ \cline{2-4} 
 &  & {\color[HTML]{0475C1} LAPQ} (Ours) & \textbf{75.1} \\ 
 &  & {\color[HTML]{FF7C37} DUAL}  \cite{choukroun2019low} & 73.06 \\ 
 &  & {\color[HTML]{3F7E31} ACIQ}  \cite{banner2018post} & 66.42 \\  & \multirow{-4}{*}{8 / 4} & {\color[HTML]{999903} MMSE} & 74.3 \\\cline{2-4} 
 &  & {\color[HTML]{0475C1} LAPQ} (Ours) & \textbf{64.4} \\ 
 &  & {\color[HTML]{3F7E31} ACIQ}  \cite{banner2018post} & 31.01 \\ & \multirow{-3}{*}{8 / 3} & {\color[HTML]{999903} MMSE} & 54.1 \\
 \cline{2-4} 
 &  & {\color[HTML]{0475C1} LAPQ} (Ours) & \textbf{38.6} \\ 
   &  & {\color[HTML]{3F7E31} ACIQ}  \cite{banner2018post} & 0.5 \\ 

 &  & {\color[HTML]{9A0000} KLD}  \cite{migacz20178} & 1.84 \\ 
\multirow{-12}{*}{Inception-V3} & \multirow{-4}{*}{4 / 4} & {\color[HTML]{999903} MMSE} & 2.2 \\ \bottomrule
\end{tabular}
}
\end{subtable}
\end{table}

We evaluate our method on several CNN architectures on ImageNet. We select a calibration set of 512 random images for the optimization step. The size of the calibration set defines the trade-off between generalization and running time (the analysis is given in \cref{app:calibration}).  %Then take clipping values found by our method and check on validation set.  
Following the convention \cite{baskin2018nice,yang2019quantization}, we do not quantize the first and last layers. 

Many successful methods of post-training quantization perform finer parameter assignment, such as group-wise \cite{Mellempudi2017ternary}, channel-wise \cite{banner2018post}, pixel-wise \cite{faraone2018syq} or filter-wise \cite{choukroun2019low} quantization, which require special hardware support and additional computational resources. Finer parameter assignment appears to provide unconditional improvement, independently of the underlying methods used. In contrast with those approaches, our method performs layer-wise quantization, which is simple to implement on any existing hardware that supports low precision integer operations. Thus, we do not include the above-mentioned methods in our comparison study. We apply bias correction, as proposed by \citet{banner2018post}, on top of the proposed method.
In \cref{tab:model_comparison} (more results are available in \cref{tab:model_comparison2} in the Appendix) we compare our method with several other layer-wise quantization methods, as well as the minimal MSE baseline. In most cases, our method significantly outperforms all the competing methods, showing acceptable performance even for 4-bit quantization.

\subsection{NCF-1B}
In addition to the vision models, we evaluated our method on a recommendation system task, specifically on a Neural Collaborative Filtering (NCF) \cite{Xiangnan2017ncf} model. We use mlperf\footnote{\url{https://github.com/mlperf/training/tree/master/recommendation/pytorch}} implementation to train the model on the MovieLens-1B dataset. Similarly to the ImageNet, the calibration set of 50k random user/item pairs is significantly smaller than both the training and validation sets. 
% . Calculate optimal clipping values on the calibration set and then check performance on validation set of 20M examples.

\begin{table}
\centering
\caption{Hit rate of NCF-1B applying our method, {\color[HTML]{0475C1} LAPQ}, and {\color[HTML]{999903} MMSE}.
}
\label{tab:ncf}
\begin{tabular}{lccc}
\hline
\textbf{Model}                   & \textbf{W/A}                   & \textbf{Method} & \textbf{Hit rate(\%)} \\ \toprule
\multirow{7}{*}{NCF 1B} & 32/32                 & FP32   & 51.5   \\ \cline{2-4} 
                        & \multirow{2}{*}{32/8} & {\color[HTML]{0475C1} LAPQ} (Ours)  & \textbf{51.2}   \\ 
                        &                       & {\color[HTML]{999903} MMSE}   & 51.1   \\ \cline{2-4} 
                        & \multirow{2}{*}{8/32} & {\color[HTML]{0475C1} LAPQ} (Ours)   & \textbf{51.4}   \\
                        &                       & {\color[HTML]{999903} MMSE}   & 33.4   \\ \cline{2-4} 
                        & \multirow{3}{*}{8/8}  & {\color[HTML]{0475C1} LAPQ} (Ours)   & \textbf{51.0}     \\ 
                        &                       & {\color[HTML]{999903} MMSE}   & 33.5 \\ \bottomrule
\end{tabular}
\end{table}

In \cref{tab:ncf} we present results for the NCF-1B model compared to the MMSE method. Even at 8-bit quantization, NCF-1B suffers from significant degradation when using the naive MMSE method. In contrast, LAPQ achieves near baseline accuracy with 0.5\% degradation from FP32 results.

%\subsection{Pascal VOC}
%We evaluated our method on detection task, specifically on the Pascal VOC object detection challenge \cite{pascalvoc} with \BCH{explain the architecture and cite + explain results}.

%\BCH{Yury will add results for detection task}
%\CB{The same model as in
%\cite{fang2020nearlossless}}

\subsection{Ablation study}
\label{sec:ablation}

% \paragraph {Coordinate descent}
% Having a function $\mathcal{F}$ of $N$ parameters $C = \qty{c_1, c_2, \dots, c_N }$, coordinate descent optimize each parameter independently from the others. 
% Given minimization problem:
% \begin{equation}
%     \min_{c \in C}\mathcal{F}\qty(c_1,c_2, \dots , c_N)
% \end{equation}
% we optimize one parameter while fix the others. Coordinate descent is a very simple and scalable method of optimization, however it does not provides any convergence guarantees. In case of separable functions, however, if single-variable optimization achieves minimum, so does coordinate descent.

\paragraph {Initialization}
The proposed method comprises of two steps: layer-wise optimization and quadratic approximation as the initialization for the joint optimization. In \cref{tab:abl-init}, we show the results for ResNet-18 under different initializations and their results after adding the joint optimization. LAPQ suggest a better initialization for the joint optimization.

\begin{table}
\centering
\caption{Accuracy of ResNet-18 with three different initializations. LW  refers to only applying layer-wise optimization for $p=2$. LW + QA refers to the proposed initialization that includes layer-wise optimization and quadratic approximation. On top of each initialization, we apply the joint optimization.  }
\label{tab:abl-init}
\begin{tabular}{llcc}
\toprule
\multirow{2}{*}{\textbf{W / A}} & \multirow{2}{*}{\textbf{Method}} & \multicolumn{2}{c}{\textbf{Accuracy (\%)}}   \\ \cline{3-4}
 &  & \textbf{Initial} & \textbf{Joint}   \\ \midrule
\multirow{3}{*}{4 / 4} & Random & 0.1 & 7.8   \\ 
 & LW & 43.6 & 57.8   \\ 
 & LW + QA & 54.1 & 60.3   \\ \midrule
\multirow{3}{*}{32 / 2} & Random & 0.1 & 0.1   \\ 
 & LW & 33 & 50.3   \\
 & LW + QA & 48.1 & 50.7   \\ \bottomrule
\end{tabular}
\end{table}

\paragraph{Bias correction}

Prior research \cite{banner2018post,finkelstein2019fighting} has shown that CNNs are sensitive to quantization bias. To address this issue, we perform bias correction of the weights as proposed by \citet{banner2018post}, which can easily be combined with LAPQ, in all our CNN experiments. In \cref{tab:bias} we show the effect of adding bias correction to the proposed method and compare it with MMSE.  We see that bias correction is especially important in compact models, such as MobileNet.

\begin{table}
\centering
\caption{Effect of applying bias correction on top of LAPQ for ResNet-18, ResNet-50 and MobileNet-V2. LAPQ significantly outperforms naive minimization of the mean square error. Notice the importance of bias correction on MobileNet-V2. }
\begin{tabular}{ccccc}
\toprule
W            & A           & ResNet-18 & ResNet-50 & MobileNet-V2 \\ \toprule
\multicolumn{5}{c}{LAPQ}                                      \\ \midrule
32           & 4           & \textbf{68.8\%}   & \textbf{74.8\%}   & \textbf{65.1\%}       \\ 
32           & 2           & \textbf{51.6\%}   & \textbf{54.2\%}   & \textbf{1.5\%}        \\ 
4            & 32          & 62.6\%   & 69.9\%   & 29.4\%       \\ 
4            & 4           & 58.5\%   & 66.6\%   & 21.3\%       \\ \toprule
\multicolumn{5}{c}{LAPQ + bias correction}                    \\ \midrule
4            & 32          & \textbf{63.3\%}   & \textbf{71.8\%}   & \textbf{60.2\%}       \\
4            & 4           & \textbf{59.8\%}   & \textbf{70.\%}    & \textbf{49.7\%}       \\ \toprule
\multicolumn{2}{c}{FP32} & 69.7\%   & 76.1\%   & 71.8\%       \\ \bottomrule
\end{tabular}
\label{tab:bias}
\end{table}

% \cref{tab:models_ablation} report accuracy at different bitwidths compared to minimal MSE baseline. Our method provides better improvements over baseline at lower bitwidth. Moreover, we note that weights are more sensitive to quantization than activations. Even at 4 bit quantization, minimization of MSE results in significant accuracy degradation compared to LAPQ.  As shown in \cref{tab:models_ablation}, bias correction significantly reduces  quantization error and improves  accuracy of the MobileNet as well as other networks.

%% file: sections/060_conclusion.tex
\section{Conclusion}
\label{sec:conclusion}
We have analyzed the loss function of quantized neural networks. At low precision, the function is non-separable, with steep curvature, which is unfavorable for existing post-training quantization methods. Accordingly, we have introduced Loss Aware Post-training Quantization (LAPQ), which jointly optimizes all quantization parameters by minimizing the loss function directly. We have shown that our method outperforms current post-training quantization methods. 
Also, our method does not require special hardware support such as channel-wise or filter-wise quantization. 
In some models, LAPQ is the first to show compatible performance in 4-bit post-training quantization regime with layer-wise quantization, almost achieving a the full-precision baseline accuracy.

%% file: sections/070_appendix.tex
\section{Hessian of the loss function}

To estimate dependencies between clipping parameters of different layers, %and inside each layer 
we analyze the structure of the Hessian matrix of the loss function. The Hessian matrix contains the second-order partial derivatives of the loss $\mathcal{L}(\Delta)$, where $\vb{\Delta}$ is a vector of quantization steps: 
\begin{align}
    H[\mathcal{L}]_{ij} = \pdv{\mathcal{L}}{\Delta_i}{\Delta_j}.
\end{align}

\begin{figure*}
\begin{subfigure}{0.49\linewidth}
    \includegraphics[width=\linewidth]{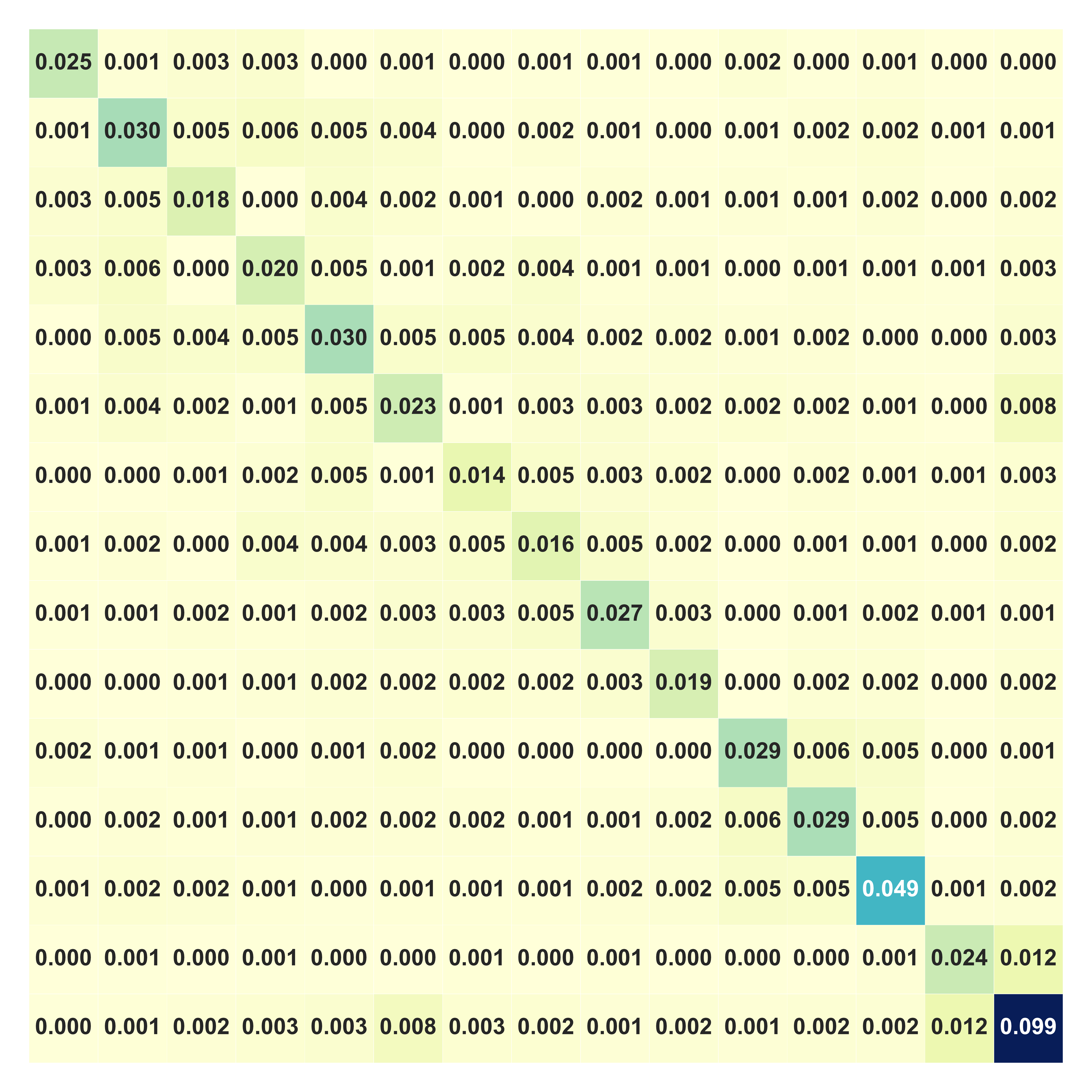}
    \subcaption{Quantization of 4 bit}
\end{subfigure}
\hfill
\begin{subfigure}{0.49\linewidth}
    \includegraphics[width=\linewidth]{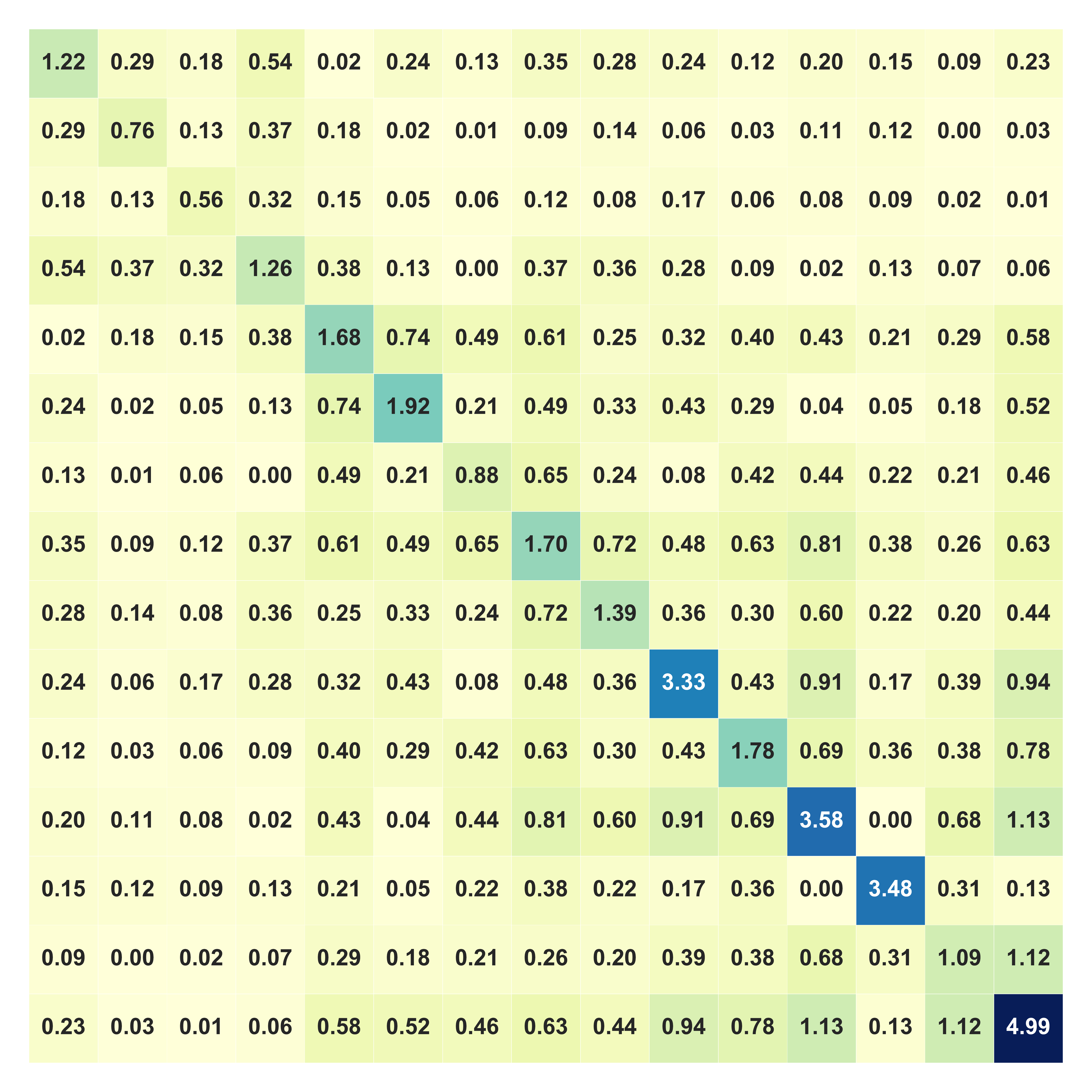}
    \subcaption{Quantization of 2 bit}
\end{subfigure}
\caption{ Absolute value of the Hessian matrix of the loss function with respect to quantization steps calculated over 15 layers of ResNet-18. Higher values at a diagonal of the Hessian at 2-bit quantization suggest that the minimum is sharper than at 4 bits. Non-diagonal elements provide an indication of the coupling between parameters of different layers: closer layers generally exhibit stronger interactions.}
\label{fig:hessian}
\end{figure*}

In the case of separable functions, the Hessian is a diagonal matrix. This means that the magnitude of the off-diagonal elements  can be used as a measure of separability.
In \cref{fig:hessian} we show the Hessian matrix of the loss function under quantization of 4 and 2 bits. As expected, higher dependencies between quantization steps emerge under more aggressive quantization.

\section{Calibration set size}
 \label{app:calibration}
\begin{figure}
\centering
    \includegraphics[width=0.6\linewidth]{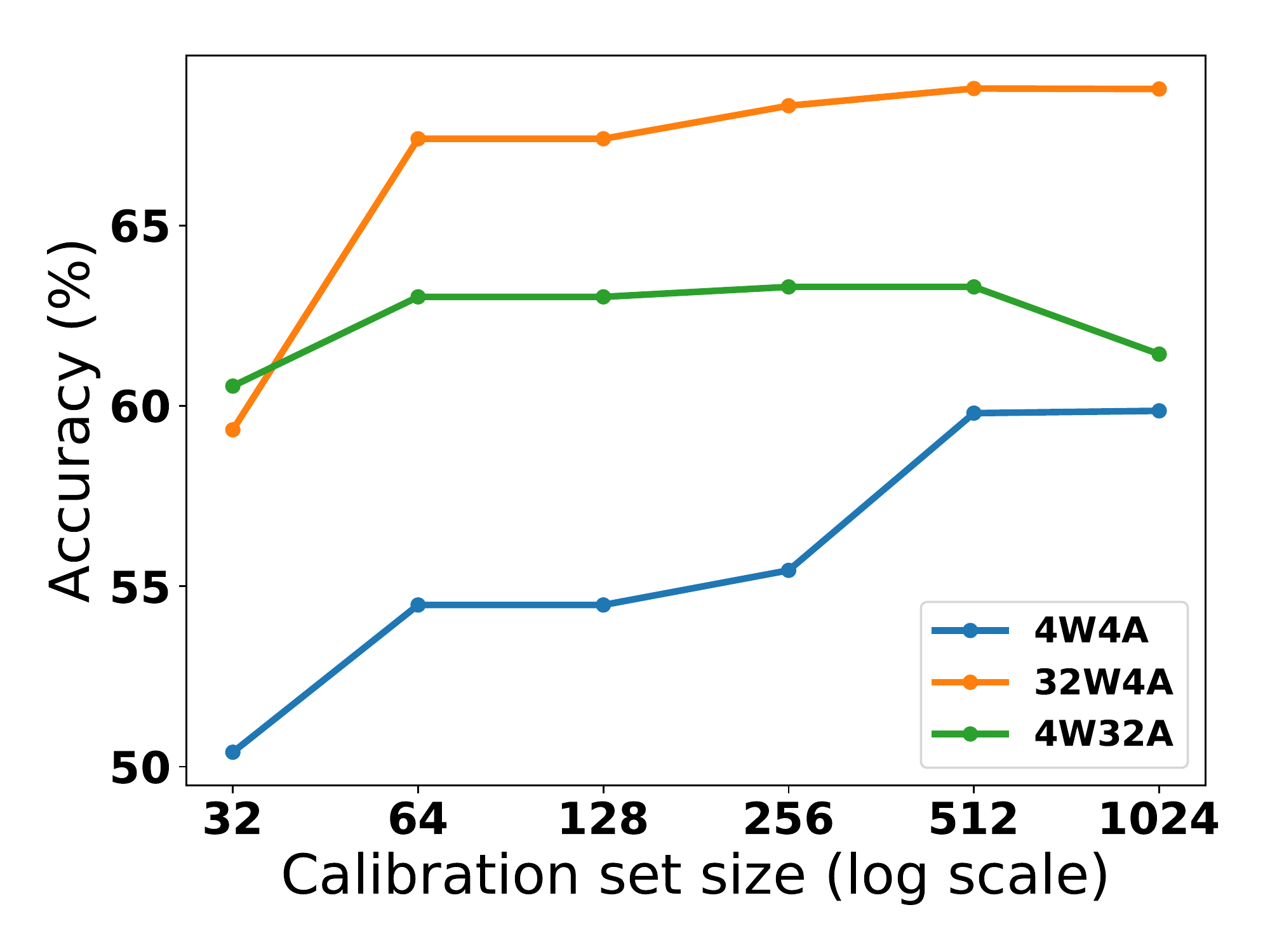}
    \caption{Accuracy of ResNet-18 for different sized calibration sets at various quantization levels.}
    \label{fig:res18_CalibrationVsAcc}
\end{figure}

Calibration set size reflects the balance between the running time and the generalization. To determine the required size, we ran the proposed method on ResNet-18 for various calibration set sizes and different bitwidths.
As shown in \cref{fig:res18_CalibrationVsAcc}, a calibration set size of 512 is a good choice to balance this trade-off.

\section{ImageNet additional results}
In \cref{tab:model_comparison2} we show additional results for the proposed method for the image classification task on ImageNet dataset.

\begin{table}
\centering
\caption{Comparison with other methods.}
\label{tab:model_comparison2}
\begin{tabular} {lclc}
\toprule
\textbf{Model} & \textbf{W/A} & \textbf{Method} & \textbf{Accuracy(\%)} \\ \midrule
 & 32 / 32 & FP32 & 69.7 \\ \cline{2-4} 
\multirow{-1}{*}{ResNet-18} & \multirow{2}{*}{8 / 2} &    {\color[HTML]{0475C1} LAPQ} (Ours) & \textbf{51.6} \\ 
& & {\color[HTML]{3F7E31} ACIQ}  \cite{banner2018post} & 7.07 \\ \midrule
 & 32 / 32 & FP32 & 76.1 \\ \cline{2-4} 
 &  & {\color[HTML]{0475C1} LAPQ} (Ours) & \textbf{54.2} \\ 
 & \multirow{-2}{*}{8 / 2} & {\color[HTML]{3F7E31} ACIQ}  \cite{banner2018post} & 2.92 \\ \cline{2-4} 
 &  & {\color[HTML]{0475C1} LAPQ} (Ours) & \textbf{71.8} \\ 
\multirow{-5}{*}{ResNet-50}  & \multirow{-2}{*}{4 / 32} & {\color[HTML]{872996} OCS} \cite{zhao2019improving} & 69.3 \\ 
   \midrule
 & 32 / 32 & FP32 & \multicolumn{1}{c}{77.3} \\ \cline{2-4} 
 &  & {\color[HTML]{0475C1} LAPQ} (Ours) & \textbf{29.8} \\ 
 & \multirow{-2}{*}{8 / 2} & {\color[HTML]{3F7E31} ACIQ}  \cite{banner2018post} & 3.826 \\ \cline{2-4} 
 &  & {\color[HTML]{0475C1} LAPQ} (Ours) & \textbf{66.5} \\ 
 \multirow{-5}{*}{ResNet-101} & \multirow{-2}{*}{4 / 32} & {\color[HTML]{999903} MMSE} & 18 \\ 
 \midrule
%   & 32 / 32 & FP32 & \multicolumn{1}{c}{77.2} \\ \cline{2-4} 
%   &  & {\color[HTML]{0475C1} LAPQ} (Ours) & \textbf{74.4} \\ 
%   &  & {\color[HTML]{FF7C37} DUAL}  \cite{choukroun2019low} & 73.06 \\ 
%   &  & {\color[HTML]{872996} OCS} \cite{zhao2019improving} & 0.2 \\  &  & {\color[HTML]{0475C1} LAPQ} (Ours) & \textbf{51.6} \\ 
%   & \multirow{-2}{*}{4 / 32} & {\color[HTML]{999903} MMSE} & 5.8 \\ \cline{2-4} 
%  \multirow{-12}{*}{Inception-V3} & \multirow{-3}{*}{4 / 4} &  \bottomrule
\end{tabular}

\end{table}